\documentclass{article}

\usepackage{PRIMEarxiv}

\usepackage[utf8]{inputenc} 
\usepackage[T1]{fontenc}    
\usepackage{hyperref}       
\usepackage{url}            
\usepackage{booktabs}       
\usepackage{amsfonts}       
\usepackage{nicefrac}       
\usepackage{microtype}      
\usepackage{lipsum}
\usepackage{fancyhdr}       
\usepackage{graphicx}       
\graphicspath{{media/}}     
\usepackage{amsmath}
\usepackage{array} 
\usepackage{multirow}
\title{SAM-IF: Leveraging SAM for Incremental Few-Shot Instance Segmentation
}

\author{
  Xudong Zhou \\
  Shanghai Jiao Tong University \\
  Shanghai\\
  \texttt{dong2002@sjtu.edu.cn} \\
   \And
  Wenhao He \\
  Noematrix Co., Ltd. \\
  Shanghai\\
  \texttt{wenhao.he@noematrix.cn} \\
}

\begin{document}
\maketitle

\begin{abstract}
We propose SAM-IF, a novel method for incremental few-shot instance segmentation leveraging the Segment Anything Model (SAM). SAM-IF addresses the challenges of class-agnostic instance segmentation by introducing a multi-class classifier and fine-tuning SAM to focus on specific target objects. To enhance few-shot learning capabilities, SAM-IF employs a cosine-similarity-based classifier, enabling efficient adaptation to novel classes with minimal data. Additionally, SAM-IF supports incremental learning by updating classifier weights without retraining the decoder. Our method achieves competitive but more reasonable results compared to existing approaches, particularly in scenarios requiring specific object segmentation with limited labeled data.
\end{abstract}

\section{Introduction}

In recent years, class-agnostic instance-level segmentation has emerged as a critical task, where the goal is to segment instances of interest in an image without relying on predefined object classes. Segment Anything Model (SAM)\cite{kirillov2023sam} has shown potential for various segmentation tasks due to its flexible prompting mechanism, which includes point-based, bounding-box-based, and global content segmentation using point grids. However, users do not always have access to predefined points or bounding boxes, and may only require masks for specific objects of interest. To address this need, we leverage datasets with annotated masks for fine-tuning SAM to selectively identify and segment target objects.

In its original form, SAM’s adapter allows for training on single-mask images but lacks support for instance-level segmentation in scenarios involving multiple objects, each with separate masks. To overcome this limitation, we have developed an instance-level fine-tuning approach for SAM, introducing a multi-class classifier including a background class to ignore irrelevant parts of the scene, such as background objects (e.g., cups, shelves). Our approach ensures that the fine-tuned SAM model produces meaningful segmentations without extraneous background regions. Additionally, we integrate the improved SAM2 \cite{ravi2024sam2} model, which offers enhanced performance and efficiency, further refining SAM’s ability to handle instance-level segmentation and foreground-background separation.

To extend the functionality of our model, we explore few-shot learning for multi-class instance-level segmentation, building upon SAM’s strong generalization capabilities. Inspired by the incremental few-shot instance segmentation approach in iMTFA ~\cite{zhang2020imtfa}, we first train SAM's decoder on the COCO2014 dataset, optimizing the segmentation results and suppressing background segments. We then train a cosine-similarity-based classifier on a base dataset of 60 classes, without updating the decoder, and calculate cosine similarities for each novel class sample to insert similarity weights into the classifier. This enables the model to achieve accurate segmentation on novel classes with few-shot samples.

Our contributions can be summarized as follows:
\begin{itemize}
    \item We present the first instance segmentation approach based on SAM, enabling instance-level segmentation and selective multi-class classification.
    \item We introduce the first incremental few-shot learning framework based on SAM, demonstrating effective multi-class instance segmentation with limited samples.
\end{itemize}

\section{Related Work}

\subsection{Segment Anything Model (SAM)}
The Segment Anything Model (SAM) \cite{kirillov2023sam} introduced by Meta AI provides a flexible, prompt-based interface for segmentation tasks, supporting points, bounding boxes, and grid-based global segmentation. However, SAM has notable limitations, including high computational demands and restricted prompt flexibility, which hinder its applicability in certain scenarios. Specifically, SAM's existing mechanisms are insufficient for class-agnostic instance segmentation, particularly in cases where users lack access to point or bounding box prompts and only need segmentation of specific objects.

SAM2 \cite{ravi2024sam2} addresses SAM's computational overhead issues by improving memory efficiency and inference speed, making it more practical for end-to-end fine-tuning. In our work, we leverage SAM2 to perform fine-tuning for class-agnostic instance segmentation. This is achieved by training on selective masks and incorporating a multi-class classifier including a background class, which allows us to extend SAM2’s capabilities and apply it to scenarios where class-specific labels are unavailable, enabling efficient and adaptable segmentation solutions.

\subsection{Few-Shot Learning in Segmentation}

Few-shot learning (FSL) in segmentation enables models to adapt to new, unseen object categories with limited labeled data. Early works in FSL for segmentation focused on using metric learning and meta-learning techniques to generalize across tasks with minimal examples~\cite{vinyals2016matching, ravi2017optimization, shaban2017oneshot, snell2017prototypical, bertinetto2016meta}. In particular, one-shot learning frameworks for semantic segmentation have been proposed~\cite{shaban2017oneshot}, demonstrating that models could be adapted to segment novel objects based on a single labeled example. Other significant contributions in meta-learning include the works of Munkhdalai and Yu~\cite{munkhdalai2017meta}, which provide foundational approaches for task adaptation in few-shot learning, and Lake et al.~\cite{lake2015human}, who introduced probabilistic program induction to support human-level concept learning.

One significant advance in segmentation was the development of Mask R-CNN~\cite{he2017mask}, which extended Faster R-CNN to generate high-quality object masks for each region. This method, alongside other deep learning-based segmentation approaches, has set the foundation for few-shot segmentation. Recent works have introduced techniques for incremental learning in few-shot segmentation, enabling models to handle new object classes without retraining from scratch. These methods not only reduce memory requirements compared to traditional approaches like FGN~\cite{shaban2017oneshot} and Siamese Mask R-CNN~\cite{koch2018siamese}, but also allow for incremental class addition, making them highly efficient for dynamic environments.

The integration of attention mechanisms and generative models~\cite{liu2018learning, tzeng2019further} has further enhanced the scalability and adaptability of these methods, allowing models to better handle evolving data distributions. These methods have been extended with architectures like DeepLab~\cite{chen2017multi}, which uses dilated convolutions for improved feature extraction, and Network-in-Network (NiN) for better semantic segmentation~\cite{zhou2017network}. The ability to incorporate new classes with minimal labeled data has profound implications for real-world applications, such as autonomous driving~\cite{chen2017multi} and medical imaging~\cite{zhou2017network}. The combination of metric learning, meta-learning, and incremental learning provides a robust framework for addressing the challenges posed by few-shot segmentation tasks.

\subsection{Incremental Few-shot Instance Segmentation}

Incremental learning in segmentation addresses the challenge of adapting models to new classes without retraining from scratch. Memory-based methods, such as those proposed by Cermelli et al.~\cite{cermelli2020incremental}, help prevent catastrophic forgetting by storing a subset of previous data. Meanwhile, selective learning strategies, like those in Wu et al.~\cite{wu2020incremental}, update only parts of the model relevant to new classes, improving efficiency. To handle class imbalance, Zhao et al.~\cite{zhao2018incremental} introduced a weighted loss function to prioritize new classes during training. These methods are vital for applications like autonomous driving~\cite{chen2017multi}, where new objects need to be continuously incorporated.

The iMTFA approach~\cite{zhang2020imtfa} is the first to target incremental few-shot instance segmentation (FSIS). Unlike FGN and Siamese Mask R-CNN, which require examples of every class at test time and use large amounts of memory, iMTFA can incrementally add new classes without retraining or requiring examples of base classes. Although Meta R-CNN can pre-compute per-class attention vectors, it needs retraining to accommodate a different number of classes. iMTFA's ability to add classes incrementally, without the need for extensive memory or retraining, makes it highly efficient for handling dynamic environments with limited labeled data.

\section{Methodology}

\subsection{Overview of the Model Architecture}
\begin{figure}
    \centering
    \includegraphics[width=0.9\linewidth]{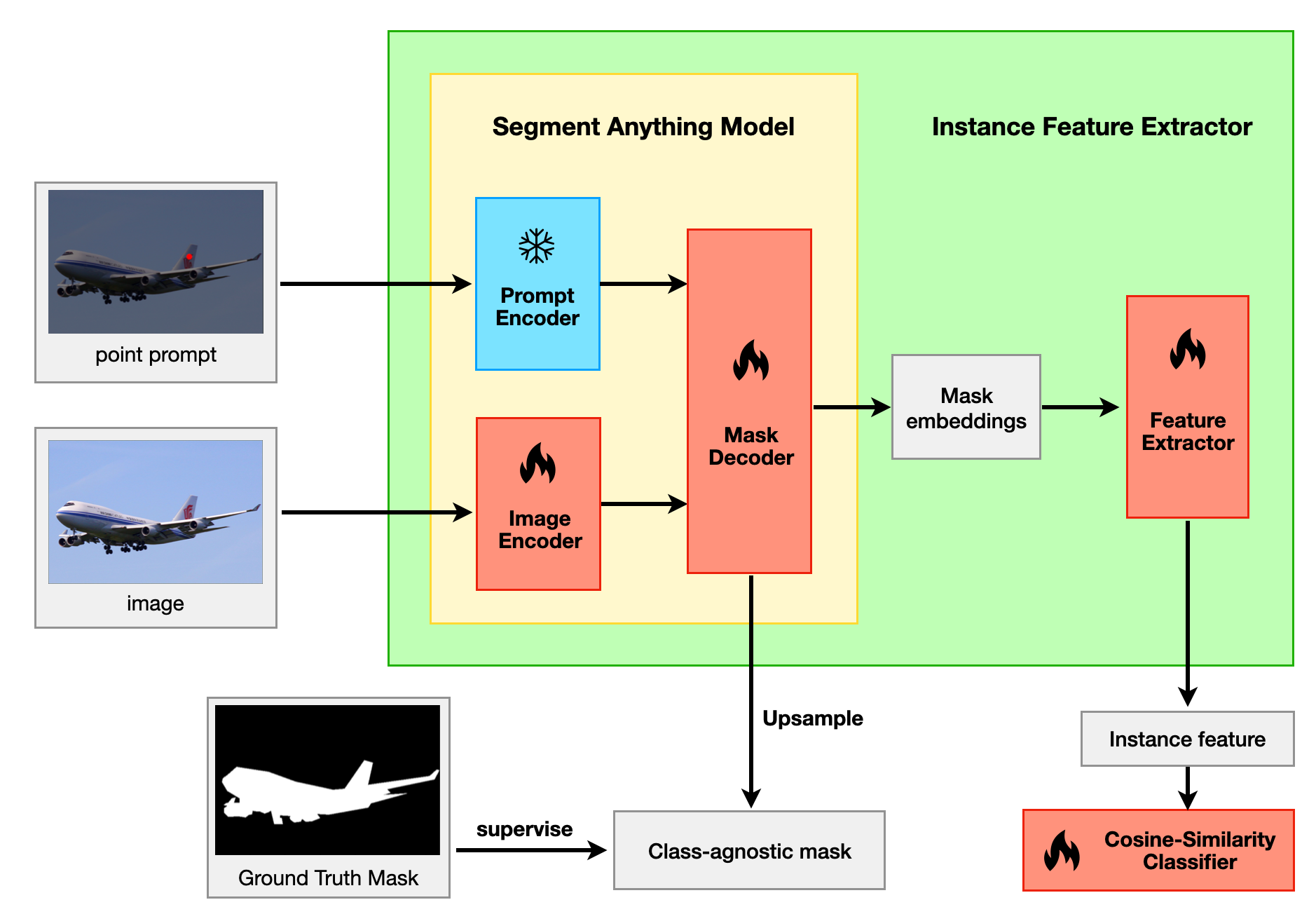}
    \caption{The architecture of SAM-IF.}
    \label{fig:arch}
\end{figure}


The architecture of SAM-IF is shown in Fig. \ref{fig:arch}, where we use the Segment Anything Model (SAM) as the mask segmenter. Initially, the points act as a sparse input, both of which are provided to the Prompt Encoder of SAM for constructing prompt features. The Prompt Encoder is frozen and not involved in the training process. Next, the image input is processed by the Image Encoder, which can be trained to improve the quality of the mask. Finally, the image and cue features are jointly fed into the mask decoder. The mask encoder produces a low-resolution mask, which is then refined through post-processing and up-sampling to generate a class-agnostic mask. Additionally, the mask embeddings are used as inputs to the multi-class classifier, which is trained from scratch. The final layer of the classifier uses a cosine similarity classification head, which calculates the relative distances between embeddings using the cosine function for classification. This approach streamlines the subsequent few-shot learning process by allowing it to skip the training phase and directly perform cosine similarity comparison and classification using the extracted features.

\paragraph{Leverage SAM2 for Segmentation}

SAM2 significantly improves upon SAM in terms of memory usage and inference speed. SAM required substantial resources and time to train the encoder and decoder, due to the large dataset and complex model architecture. SAM2, however, introduces optimizations like sparse point prompts, which reduce the need for pixel-wise labeling and lower the computational load. This results in faster inference and reduced memory consumption while maintaining competitive accuracy.

For our experiments, which involve training both encoder and decoder for class-agnostic instance-level segmentation, SAM2's efficiency is crucial. It allows us to conduct experiments with lower computational costs, without sacrificing performance. Additionally, fine-tuning SAM2 with random point prompts targets specific regions of interest, further enhancing efficiency and reducing unnecessary overhead.

\paragraph{Class-Agnostic Mask Segmentation.}

SAM2 provides sparse point prompts for inference. During the inference phase, SAM2 uniformly distributes points over the target image, evaluates the output for each point, and generates corresponding segmentation results. However, the original SAM2 model generates segmentation outputs for all objects present in the image, which may not be ideal when the goal is to focus only on specific objects within the scene.

In our approach, we perform fine-tuning by introducing random point prompts. Specifically, we randomly select points within the target image as sparse point prompts. In our method, all randomly selected points are designated as foreground during training, allowing the model to perform class-agnostic instance segmentation. In the subsequent classifier, we train the model to distinguish between foreground and background categories. This distinction serves as a basis for ignoring the background class loss in the later stages, ensuring that background segmentation is not trained. The reason for this is that foreground-background classification is unknown during inference, so when training the mask segmentation model, the focus is on the completeness and accuracy of each mask, without needing to handle background segmentation.

In the inference phase, SAM2 uniformly distributes points over the target image and generates corresponding masks based on these points. This process does not rely on prior object categories but focuses on the foreground areas within the image. To refine the segmentation, a classifier is trained to distinguish between foreground and background. The classifier’s output is used to ignore the background during inference, ensuring that the model focuses solely on instance segmentation of the target objects. For each predicted mask, a stability score is calculated to assess its reliability. A higher stability score indicates better distinction for the current category. By setting a threshold, unstable masks are filtered out, enhancing the segmentation accuracy. To eliminate redundant detection results, especially in multi-instance segmentation, Non-Maximum Suppression (NMS) is applied to remove masks with high overlap, keeping only the most discriminative ones.

\paragraph{Classifier for Point-Level Evaluation.}

We integrate a classifier to evaluate the relevance of each point in the context of the target segmentation task. The classifier uses the mask embeddings as input to classify the objects corresponding to the segmented masks, in a manner similar to iMTFA \cite{zhang2020imtfa}, by employing cosine similarity, which is computed as follows:

The classifier consists of two main components: the \textbf{Feature Extractor} and the \textbf{Cosine Similarity Layer}. 

The \textbf{Feature Extractor}, is responsible for extracting discriminative features from the input tensor \( x \). It consists of two convolutional layers (\( \text{Conv2d} \)), followed by ReLU activations (\( \text{ReLU} \)), adaptive average pooling (\( \text{AdaptiveAvgPool2d} \)), and a fully connected layer. These operations transform the input tensor into a feature vector \( \mathbf{f} \), which is then flattened into a 1D vector.

After feature extraction, we compute the cosine similarity between the extracted feature vector \( \mathbf{f} \) and the weight matrix \( \mathbf{W} \) from the \textbf{Cosine Similarity Layer}. To ensure meaningful comparison, both the feature vector and the weight matrix are normalized. Specifically, the feature vector \( \mathbf{f} \) is normalized by dividing it by its L2 norm \( \|\mathbf{f}\| \), and similarly, the weight matrix \( \mathbf{W} \) is normalized by dividing it by its L2 norm \( \|\mathbf{W}\| \). The cosine similarity is computed as the dot product of the normalized feature vector and the normalized weights, followed by a scaling factor \( \gamma \), which can be written as:

\[
\text{cos\_sim\_pred} = \left( \frac{\mathbf{f}}{\|\mathbf{f}\|} \cdot \frac{\mathbf{W}}{\|\mathbf{W}\|} \right) \cdot \gamma
\]

where \( \gamma \) is a scaling factor (in this case, \( \gamma = 7 \)).

Finally, the predicted class scores \( \mathbf{y} \) are computed as:

\[
\mathbf{y} = \gamma \cdot \left( \frac{\mathcal{F}(x) \cdot \mathbf{W}}{\|\mathcal{F}(x)\| \|\mathbf{W}\|} \right)
\]

This expression represents the complete forward pass of the network, where \( \mathcal{F}(x) \) is the output of the Feature Extractor and \( \mathbf{W} \) is the weight matrix from the Cosine Similarity Layer.

\subsection{Incremental Few-Shot Learning with Cosine-Similarity}

\begin{figure}
    \centering
    \includegraphics[width=0.75\linewidth]{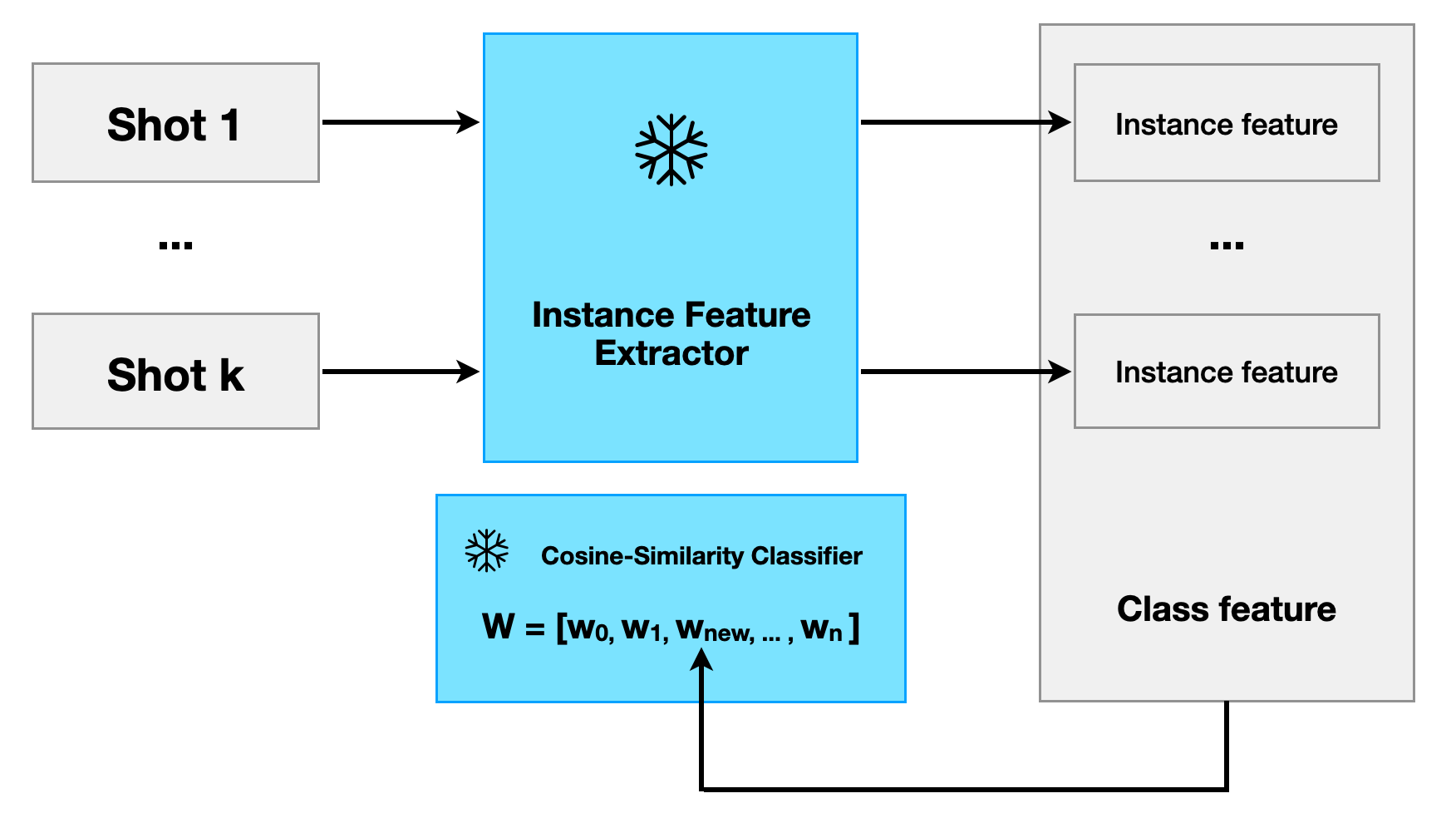}
    \caption{Construction of Class Weights for Novel Categories.}
    \label{fig:incremental}
\end{figure}

In the context of incremental few-shot learning, our goal is to update the class weights for novel categories without retraining the entire network. Instead, we incrementally learn the category-specific feature vectors and update the classifier’s weights by leveraging cosine similarity. 

As illustrated in Fig. \ref{fig:incremental}, we begin by extracting the mask embeddings \( \mathbf{m}_c \) for each novel category. These embeddings are then used to compute the new feature vector \( w_{\text{new}} \) for each category. The feature vector \( w_{\text{new}} \) is derived by passing the mask embeddings through the \textbf{Feature Extractor and Cosine-Similarity Classifier}, as shown in the following equation:

Once we obtain \( w_{\text{new}} \), we normalize the vector and update the class weights \( W_c \) as follows:

\[
W_c = \frac{w_{\text{new}}}{\|w_{\text{new}}\|}
\]

Next, the class weight updates are averaged over the number of shots, \( n_{\text{shots}} \), as follows:

\[
W_{\text{avg}} = \frac{1}{n_{\text{shots}}} \sum_{i=1}^{n_{\text{shots}}} \frac{w_{\text{new}}}{\|w_{\text{new}}\|}
\]

After normalizing the new feature vector and averaging the class weights over the shots, we update the classifier's weights for each novel category by directly replacing the corresponding position in the weight matrix \( W \). Finally, the predicted class score for each novel category is computed by measuring the cosine similarity between the updated class weights and the input point embeddings.

\subsection{Implementation Details}

\paragraph{Erosion.} 
Erosion is a morphological operation applied to both object and background masks to refine the regions from which points are sampled. The goal of the erosion is to reduce the boundaries of the mask, which prevents points from being sampled near the object’s edge and instead focuses on the object’s interior. The erosion operation is performed using a structuring element \( K \), which is typically a square or rectangular kernel. The erosion of a binary mask \( M \) can be expressed as:

\[
M' = M \circ K = \min\left( \sum_{(i,j) \in K} M(i,j), 1 \right)
\]

where \( M' \) is the eroded mask, \( M \) is the original mask, and \( K \) is the structuring element (kernel). This operation effectively shrinks the mask and ensures that only the pixels inside the object’s core are considered valid for point sampling. After applying the erosion, valid points are chosen from the eroded mask. If a valid mask is provided, it further refines the valid regions during the erosion process. The background is treated similarly, where the background mask is eroded and random points are sampled from the eroded background mask.

By applying erosion, the process ensures that the random points sampled during instance segmentation focus on the interior of the object and avoid boundary areas, leading to more stable and reliable segmentation results.

\paragraph{Instance-Level Random Sampling.} 

In the instance-level random sampling process, we perform random point selection within the region corresponding to each instance. Rather than using the entire image's background weight for sampling, we assign uniform weight to each instance. This ensures that even smaller objects, like cups or skateboards, which may be overwhelmed by the background's weight, have an equal chance of being sampled. This approach ensures that both small and large objects achieve good segmentation results and classification accuracy, as the weight distribution is more balanced.

\section{Experiments}

\subsection{Experiment Setup}

Our evaluation procedure follows conventions established in iMTFA \cite{zhang2020imtfa}. Specifically, we evaluate on the COCO2014 dataset and adopt a class split where the 80 COCO classes are divided into 20 novel classes and 60 base classes, with the novel classes overlapping with those in VOC. The training set consists of the union of COCO’s 80k train and 35k validation images, while the test set includes the remaining approximately 5k images.

In this work, we focus on the 1-shot setting for the novel classes. For each novel class, one random example (shot) is selected for training. To mitigate the effect of randomness in shot selection, we repeat the 1-shot evaluation process 10 times with different random samples and report the mean result. Our evaluation procedure adheres to the typical few-shot evaluation setting.

The experiments are conducted on an NVIDIA A100 80G GPU. The learning rates are set as follows: for the image encoder and mask decoder, the learning rate is $0.0001$, and for the classifier, it is $0.005$. Additionally, the parameter $\gamma$ is set to $7$. The batch size for images is set to $11$, and the batch size for random points is $16$, which nearly utilizes the full GPU memory capacity. These hyperparameter choices and configurations are consistent across all experiments to ensure fairness and reproducibility of results.

\subsection{Results Comparison}

\begin{figure}
    \centering
    \includegraphics[width=0.8\linewidth]{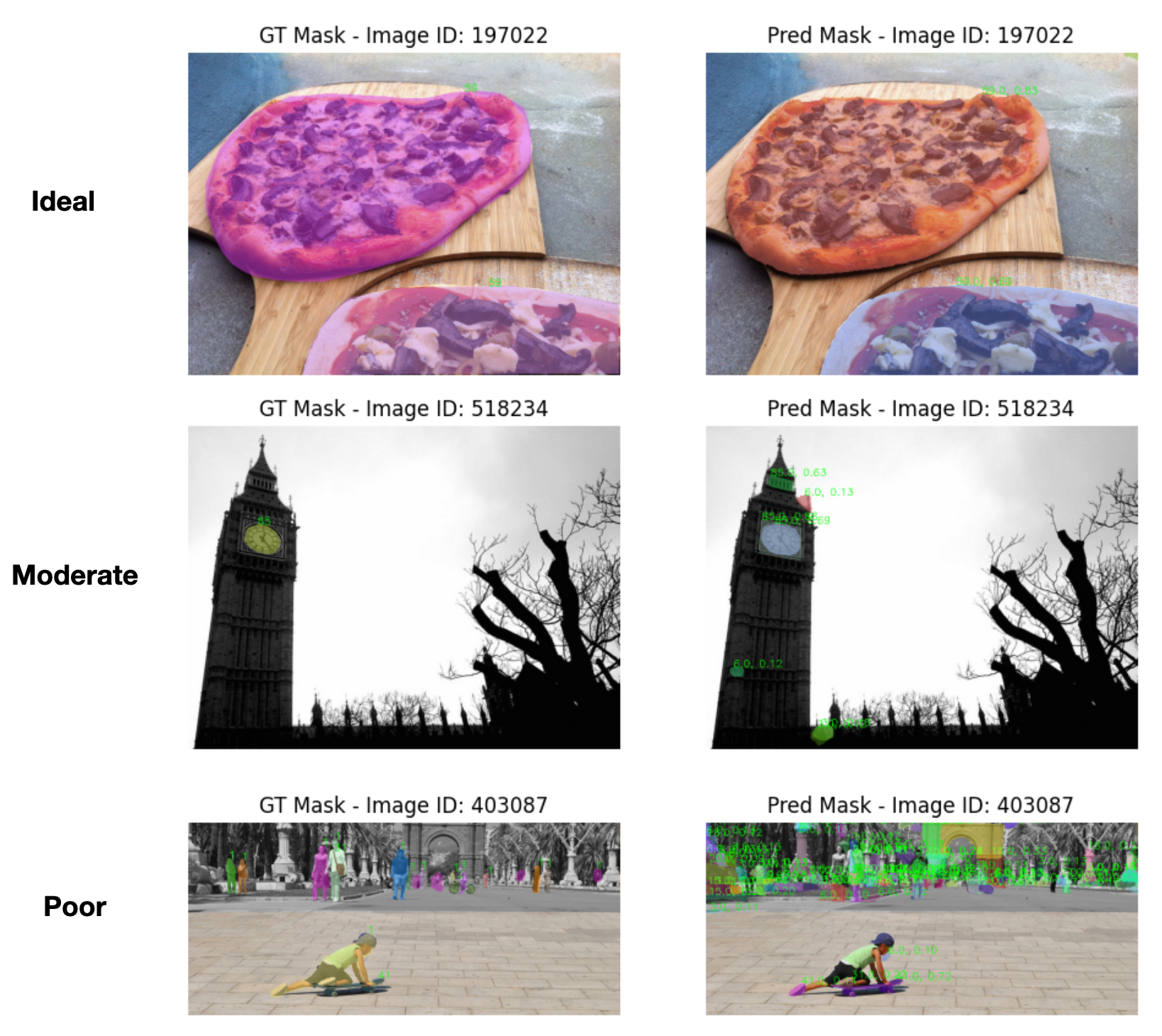}
    \caption{\textbf{Segmentation Results.} Ideal, Moderate, and Poor. The ideal result shows accurate segmentation with a clear subject and minimal background clutter. The moderate result correctly classifies the subject but includes some irrelevant segmentation. The poor result fails to segment the subject, and many small background objects are incorrectly segmented. The numbers on the left represent the class IDs, while the numbers on the right of \textit{Pred Mask} indicate the confidence scores.}
    \label{fig:segmentation_results}
\end{figure}

As shown in the Fig. \ref{fig:segmentation_results}, the ideal result demonstrates accurate segmentation, where the subject is clearly identified and the background is relatively clean with few irrelevant objects. This outcome is indicative of a well-performing model that can accurately differentiate between the subject and the background. In the moderate result, although the subject is correctly classified, the segmentation includes some irrelevant objects, highlighting a slight decrease in performance. Lastly, the poor result fails to properly segment the subject, primarily because the "person" class is not included in the base classes. Additionally, many small background objects are incorrectly segmented, which indicates a significant failure in both subject identification and background separation.

\renewcommand{\arraystretch}{1.5} 
\begin{table}[h!]
\centering
\caption{Comparison of Segmentation Metrics between iMTFA and SAM-based Methods}
\label{tab:segmentation_comparison}
\begin{tabular}{|m{1.5cm}<{\centering}|m{2.5cm}<{\centering}|m{1.5cm}<{\centering}|m{1.5cm}<{\centering}|m{1.5cm}<{\centering}|m{1.5cm}<{\centering}|m{1.5cm}<{\centering}|m{1.5cm}<{\centering}|}
\hline
\multirow{3}{*}{\textbf{Shot}} & \multirow{3}{*}{\textbf{Method}} & \multicolumn{6}{c|}{\textbf{Segmentation}}                                                                                                             \\ \cline{3-8} 
                               &                                  & \multicolumn{2}{c|}{\textbf{Overall}}                   & \multicolumn{2}{c|}{\textbf{Base}}                     & \multicolumn{2}{c|}{\textbf{Novel}} \\ \cline{3-8} 
                               &                                  & \multicolumn{1}{c|}{AP}    & \multicolumn{1}{c|}{AP50}  & \multicolumn{1}{c|}{AP}   & \multicolumn{1}{c|}{AP50}  & \multicolumn{1}{c|}{AP}     & AP50  \\ \hline
\multirow{3}{*}{1}             & iMTFA                            & \multicolumn{1}{c|}{20.13} & \multicolumn{1}{c|}{30.64} & \multicolumn{1}{c|}{25.9} & \multicolumn{1}{c|}{39.28} & \multicolumn{1}{c|}{2.81}   & 4.72  \\ \cline{2-8} 
                               & SAM-IF                           & \multicolumn{1}{c|}{17.8}  & \multicolumn{1}{c|}{27.7}  & \multicolumn{1}{c|}{18.1} & \multicolumn{1}{c|}{28.5}  & \multicolumn{1}{c|}{0.5}    & 1     \\ \cline{2-8} 
                               & SAM-IF Base                      & \multicolumn{1}{c|}{17.8}  & \multicolumn{1}{c|}{27.6}  & \multicolumn{1}{c|}{18.1} & \multicolumn{1}{c|}{28.5}  & \multicolumn{1}{c|}{-}      & -     \\ \hline
\end{tabular}
\end{table}

As shown in Table \ref{tab:segmentation_comparison}, our method demonstrates promising results in segmentation tasks when compared to iMTFA. While there is still a performance gap in specific metrics such as AP and AP50, the SAM-based methods show notable improvements in generalization. This is evident in the "Overall" metrics, where SAM methods achieve results that are closer to iMTFA. This indicates the effectiveness of SAM in enhancing generalization performance, particularly in few-shot segmentation scenarios.

\subsection{Ablation Study}
\renewcommand{\arraystretch}{1.5} 
\begin{table}[h!]
\centering
\caption{Performance Comparison of Freezing vs. Training the Image Encoder}
\label{tab:encoder_freeze_comparison}
\begin{tabular}{|m{3cm}<{\centering}|m{2.5cm}<{\centering}|m{1.5cm}<{\centering}|m{1.5cm}<{\centering}|}
\hline
\multirow{3}{*}{\textbf{Encoder Freeze}} & \multirow{3}{*}{\textbf{Method}} & \multicolumn{2}{c|}{\textbf{Segmentation}} \\ \cline{3-4} 
                                         &                                  & \multicolumn{2}{c|}{\textbf{Base}}         \\ \cline{3-4} 
                                         &                                  & \textbf{AP} & \textbf{AP50}             \\ \hline
No                                       & SAM-IF                           & 18.1        & 28.5                       \\ \hline
Yes                                      & SAM-IF                           & 14          & 21.3                       \\ \hline
\end{tabular}
\end{table}

Table \ref{tab:encoder_freeze_comparison} presents an ablation study comparing the performance of the SAM-IF model under two conditions: freezing versus training the image encoder. The results indicate that training the image encoder significantly improves the segmentation metrics, with AP increasing from 14.0 to 18.1 and AP50 rising from 21.3 to 28.5. This highlights the critical role of enabling the image encoder to update during training, as it enhances the model's ability to learn and adapt to the dataset, leading to better segmentation performance compared to freezing the encoder.

\subsection{Results Analysis}

\begin{figure}
    \centering
    \includegraphics[width=0.75\linewidth]{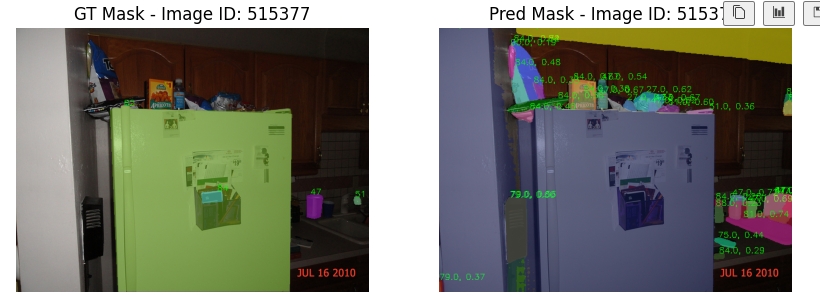}
    \caption{Analysis of low AP50 caused by missing annotations in COCO and SAM's fragmented segmentation}
    \label{fig:low-ap-analysis}
\end{figure}

The reason for the low evaluation metrics lies in the fragmented segmentation results produced by SAM and the missing annotations in COCO. Additionally, the classifier often misclassifies small objects. As shown in Fig. \ref{fig:low-ap-analysis}, the ground truth (gt) annotations only label one cup, while other cups in the image have higher confidence scores. This is reasonable; however, due to the missing annotations, the COCO evaluation algorithm significantly reduces the AP50 score. According to the COCOEval metric, this discrepancy leads to a substantial drop in the AP50 value.

\section{Future Work}
\paragraph{Enhancing Prompt Design:} Transitioning from point prompts to box or anchor prompts represents a pivotal improvement for better object localization and segmentation accuracy. This approach aligns better with COCO-style annotations and addresses the limitation of point prompts in precisely capturing object boundaries.

\paragraph{Optimizing Classifier Training:}  
An in-depth analysis of the classifier training process could help identify bottlenecks and areas for improvement. By tuning the classifier in conjunction with the encoder, we aim to enhance the synergy between segmentation and classification tasks, improving performance, especially for challenging cases such as small objects and ambiguous boundaries.

\paragraph{Reducing Classifier Dependency on SAM Embeddings:} 

SAM embeddings, while effective for segmentation, may not be ideally suited for classification tasks due to their lack of structured feature representations needed for distinguishing object classes, especially in fine-grained or novel scenarios. One possibility is to introduce a transformation layer to refine SAM embeddings, potentially improving their suitability for classification. Another approach could involve combining SAM embeddings with features from task-specific encoders, which might enhance class separability. Reducing the classifier's reliance on SAM embeddings could improve robustness, generalization to novel categories, and overall classification performance.
    
\section{Conclusion}
In this work, we proposed and implemented SAM-IF, a new approach for incremental few-shot instance segmentation built upon SAM. To enhance SAM's performance for class-agnostic segmentation, we designed a multi-class classifier including a background class and conducted fine-tuning that effectively excludes irrelevant background. By introducing a cosine-similarity-based classifier, we successfully enabled few-shot learning for novel classes, requiring only minimal labeled samples. Moreover, we achieved incremental learning by updating only the classifier weights, which allows for efficient adaptation to evolving datasets. 

Our experiments, conducted on the COCO2014 dataset, demonstrated that our method achieves competitive but more reasonable results compared to state-of-the-art approaches. These results validate SAM-IF's effectiveness in addressing challenges such as fragmented segmentation and missing annotations. In the future, we aim to improve prompt design, optimize classifier training, and reduce dependence on SAM embeddings for classification.

\section{Acknowledgments}

This work was conducted as part of an internship at Noematrix Company. We sincerely thank Noematrix for providing the resources and support that were crucial to the completion of this research.

\newpage
\bibliographystyle{unsrt}  
\bibliography{references}

\end{document}